# Tensor Graph Convolutional Network for Dynamic Graph Representation Learning

Ling Wang, Ye Yuan, *Member, IEEE*

*Abstract*—Dynamic graphs (DG) describe dynamic interactions between entities in many practical scenarios. Most existing DG representation learning models combine graph convolutional network and sequence neural network, which model spatial-temporal dependencies through two different types of neural networks. However, this hybrid design cannot well capture the spatial-temporal continuity of a DG. In this paper, we propose a tensor graph convolutional network to learn DG representations in one convolution framework based on the tensor product with the following two-fold ideas: a) representing the information of DG by tensor form; b) adopting tensor product to design a tensor graph convolutional network modeling spatial-temporal feature simultaneously. Experiments on real-world DG datasets demonstrate that our model obtains state-of-the-art performance.

## I. INTRODUCTION

Dynamic graph (DG) represents the interactions among numerous entities involved in a dynamic data-related application [1-5, 46, 50], such as social networks [6], communication networks [7], and financial transactions [8]. DG contains spatial-temporal patterns that demonstrate the behaviors of entities, which is helpful for other applications, such as planning management and decision-making [9-12]. Therefore, it is necessary to model spatial-temporal patterns for learning DG representations applied to the downstream task.

Researchers have proposed a myriad of data analysis models to extract representations from graphs [39-43]. Given a static graph, there are many models to perform representation learning. Representative ones include models based on latent features [13-15, 44, 45, 52]. When a target graph becomes dynamic, its patterns can be extracted by a tensor-based model [16-18, 47, 51]. However, the presentation learning ability of models based on matrix/tensor decomposition may be limited since they neglect the graph structure of a DG.

For representation learning by exploiting graph structure, Graph Convolutional Networks (GCN) [19] have gained attention for their ability to extend the convolutional operator from Euclidean space data to graph-structured data [20-22]. DG represents typical graph-structured data and exhibits strong temporal correlations. Therefore, performing representation learning on such graphs demands capturing this complicated evolving nature [23-27]. In DG representation learning, the most established models are to combine GCN and sequential networks. For example, Manessi *et al.* [26] propose a WD-GCN model that combines the GCN module and the Recurrent Neural Network (RNN) module. It involves a GCN module operating on each graph snapshot independently, followed by an RNN module operating on nodes independently along the timeline. Pareja *et al.* [27] propose Evolve-GCN that uses RNN to regulate the model parameters at every time slot, which promises the parameters to change over time. However, this hybrid architecture may not be powerful enough to jointly model spatial-temporal dependencies, as it adopts the learning process of two different types of neural networks separately.

To solve the above problems, we propose a Tensor Graph Convolution Network (TGCN) by extending static GCN to DGs based on tensor product, which can directly learn DG representations in one convolution framework. We first adopt tensors to represent the necessary information of a DG. To obtain the initial features of node, we encode the unique node's identity number into a one-hot tensor and embed it into a node's latent feature tensor. Then, a tensor graph convolutional network is adopted to learn node representations in one convolution framework. Specifically, Spatial-temporal dependencies are modeled completely by tensor product using the adjacent tensor and a temporal mixing matrix. To capture the different temporal dependence scales from data, we design a learnable temporal mixing value matrix. Finally, experiments on real-world DGs assess the performance of the proposed TGCN. The results demonstrate its superiority over state-of-the-art models in accurately estimating link weight values. The main contributions of this paper can be summarized as follows: a) The information of DG is presented in a tensor form and the node information is embedded to obtain the latent feature tensor for capturing initial node features; b) We propose a tensor graph convolution model based on tensor product to model spatial-temporal dependencies under one convolution framework, in which a learnable mixing matrix is adopted to capture temporal dependencies between graph snapshots; c) We conduct extensive experiments on real-world datasets to validate the proposed TGCN. Experimental results show that the TGCN can consistently improve the DG representation learning performance.

---

⋄ L. Wang is with the School of Computer Science and Technology, Chongqing University of Posts and Telecommunications, Chongqing 400065, China, and also with the Chongqing Key Laboratory of Big Data and Intelligent Computing, Chongqing Engineering Research Center of Big Data Application for Smart Cities, and Chongqing Institute of Green and Intelligent Technology, Chinese Academy of Sciences, Chongqing 400714, China (e-mail: wangling1820@126.com).

⋄ Y. Yuan is with the College of Computer and Information Science, Southwest University, Chongqing 400715, China (e-mail: yuanyekl@gmail.com).

The remainder of this paper is organized as follows. In Section II, we briefly introduce relevant problem formulation and basic tensor algebra. In Section III, we present the proposed model. The experimental results are presented in Section IV. Conclusions are presented in Section V.

## II. PRELIMINARIES

In this section, we first present the downstream task to be solved by learning DG representations. Then, we introduce the necessary preliminaries on tensor algebra.

### A. Problem Formulation

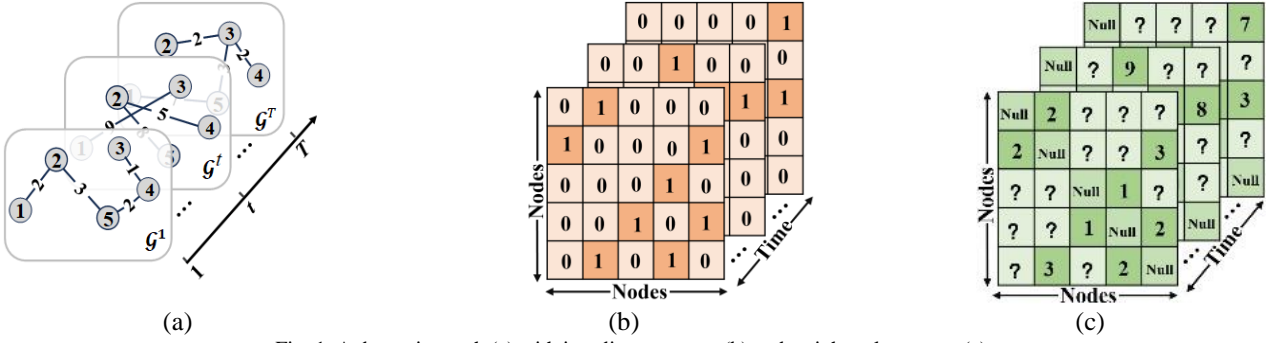

Fig. 1. A dynamic graph (a) with its adjacent tensor (b) and weight value tensor (c).

In this work, we consider the DG representation learning task. As shown in Fig. 1. (a), a DG is defined as a sequence of graph snapshots at $T$ different time slots, which is denoted as $\mathbb{G} = \{\mathcal{G}^1, \mathcal{G}^2, \ldots, \mathcal{G}^T\}$. A snapshot $\mathcal{G}^t = (\mathcal{V}, \mathcal{E}^t)$ is a graph snapshot at time slot $t$ ($1 \leq t \leq T$), where $\mathcal{V}$ is the set of all the nodes existing in the $\mathbb{G}$, and $N=|\mathcal{V}|$ is the number of nodes. $\mathcal{E}^t$ indicates the edges set of the snapshot at time $t$. The edge can be represented as an adjacency matrix $A^t \in \mathbb{R}^{N \times N}$, when there is a link between node $i$ and node $j$ at time slot $t$, i.e., $(i, j) \in \mathcal{E}^t$, $A^t_{i,j} = 1$ and otherwise $A^t_{i,j} = 0$. The link weight is captured by a weight adjacency matrix $Y^t \in \mathbb{R}^{N \times N}$, whose entry $Y^t_{i,j}$ represents the weight of the edge connecting nodes $i$ and $j$ as affected by the $A^t_{i,j}$ relation. For a DG $\mathbb{G}$ consisting of a sequence of graph snapshots, we can extend the matrix to the three-order tensor by adding the temporal dimension. Hence, the adjacency information can be presented by a tensor $\mathbf{A} \in \mathbb{R}^{N \times N \times T}$, and a weight adjacency tensor is $\mathbf{Y} \in \mathbb{R}^{N \times N \times T}$ as shown in Fig. 1 (b) and (c), respectively. Then, given a DG $\mathbb{G}$, we develop a TGCN model to extract representations of the nodes for predicting the missing link weight in the $\mathbf{Y}$.

### B. Basic Tensor Algebra

**Mode-n product**. The mode-n product generalizes the matrix-matrix product to the tensor-matrix product. Given a tensor $\mathbf{X} \in \mathbb{R}^{I_1 \times \ldots \times I_{n-1} \times I_n \times I_{n+1} \times \ldots \times I_N}$ and a matrix $U^{D \times I_n}$, then the tensor-matrix product yields a tensor $(\mathbf{X} \times_n U) \in \mathbb{R}^{I_1 \times \ldots \times I_{n-1} \times D \times \ldots \times I_N}$. Its element at position $(i_1, \ldots, i_{n-1}, d, i_{n+1}, \ldots, i_N)$ is defined as:

$$(\mathbf{X} \times_n U)_{i_1,\ldots,i_{n-1},d,i_{n+1},\ldots,i_N} = \sum_{i_n}^{I_n} U_{d,i_n} \mathbf{X}_{i_1,\ldots,i_{n-1},i_n,i_{n+1},\ldots,i_N}. \quad (1)$$

**M-transform**. Given a mixing matrix $M \in \mathbb{R}^{T \times T}$. The M-transform of a tensor $\mathbf{X} \in \mathbb{R}^{I \times J \times T}$ is denoted by $\mathbf{X} \times_3 M$ and is defined as:

$$(\mathbf{X} \times_3 M)_{i,j,t} = \sum_{k=1}^{T} M_{t,k} \mathbf{X}_{i,j,k}. \quad (2)$$

**Face-wise Product**. Given two three-order tensors $\mathbf{X} \in \mathbb{R}^{I \times J \times T}$ and $\mathbf{Y} \in \mathbb{R}^{J \times K \times T}$, the face-wise product is denoted by $(\mathbf{X} \otimes \mathbf{Y}) \in \mathbb{R}^{I \times K \times T}$ and is defined as:

$$(\mathbf{X} \otimes \mathbf{Y})_{:,:,t} = \mathbf{X}_{:,:,t} \mathbf{Y}_{:,:,t}. \quad (3)$$

**M-product**. Given two three-order tensors $\mathbf{X} \in \mathbb{R}^{I \times J \times T}$ and $\mathbf{Y} \in \mathbb{R}^{J \times K \times T}$, and an invertible matrix $M \in \mathbb{R}^{T \times T}$, the M-product is denoted by $(\mathbf{X} * \mathbf{Y}) \in \mathbb{R}^{I \times K \times T}$ and is defined as:

$$\mathbf{X} * \mathbf{Y} = \left( (\mathbf{X} \times_3 M) \otimes (\mathbf{Y} \times_3 M) \right) \times_3 M^{-1}. \quad (4)$$

## III. PROPOSED TGCN

In this paper, we propose a tensor graph convolutional network based on tensor product to exploit the temporal and spatial dependences for learning DG representations. The network consists of two main components as follows: (1) *Tensor Graph Generation*, which is adopted to embed node information to latent feature tensor and normalize adjacent tensor by tensor form. (2) *Tensor Graph Convolutional Network*, which exploits the time-varying adjacent information and the learnable temporal mixing matrix to operate tensor product on node latent feature tensor to model the spatial-temporal dependencies directly. The details of each component will be presented in the following subsections.

## A. Tensor Graph Generation

Motivating by tensor advantage, we consider constructing a tensor graph to represent the spatial and temporal dependencies in a DG completely. Hence, tensor graph generation is designed for constructing a tensor representation of DG original information to be fed into our model for subsequent calculations. Firstly, we convert the node identity number into a tensor form. For nodes in a DG, we use a one-hot tensor $\mathbf{N}$ to encode the identity of nodes in each time slot. The node $i$ at time slot $t$ is presented by vector $\mathbf{N}_{i,:,t}$, whose only the $i$-th position is 1 and all others are 0. However, the information delivered by the sparse one-hot tensor is limited. Motivated by the word2vec model [28], we adopt an embedding mechanism to map the one-hot tensor to an informative and lower-dimensional representation, where a real-valued tensor $\mathbf{X}$ is obtained to represent the latent feature tensor of nodes. A learnable matrix $W_n$ is used to embed one-hot tensor $\mathbf{N}$ by tensor mode-2 product, namely:

$$\mathbf{X} = \mathbf{N} \times_2 \mathbf{W}_n . \tag{5}$$

Then, the learnable nodes' latent feature tensor $\mathbf{X}$ is to represent the node's initial information.

The DG can be spatially represented by the adjacency tensor $\mathbf{A}$, which contains the changing link information between nodes over time. Similar to GCN using a normalized adjacent matrix [19-22], we normalized adjacent tensor by constructing a diagonal degree tensor $\mathbf{D}$ as:

$$\mathbf{D}_{i,j,k} = \begin{cases} 1 + \sum_{n=1}^{N} \mathbf{A}_{i,n,k} & i = j \\ 0 & i \neq j \end{cases} . \tag{6}$$

Then, a normalized adjacent tensor $\tilde{\mathbf{A}}$ is constructed as:

$$\mathbf{A} = \mathbf{D}^{-1/2} \otimes (\mathbf{A} + \mathbf{I}) \otimes \mathbf{D}^{-1/2}, \tag{7}$$

where $\mathbf{I}$ is a tensor, whose each frontal slice is an identity matrix.

## B. Tensor Graph Convolutional Network

The tensor graph convolutional network is inspired by the static GCN due to its effectiveness. In detail, given a static graph with adjacency matrix A and node feature matrix X, one GCN layer has the form:

$$F = \sigma(AXW), \tag{8}$$

where $\sigma$ is an activation function like Sigmoid, and W is a trainable weight matrix. Through the tensor product operation defined in (1)-(4), we can extend the dimension of the first-order GCN to the third order as TGCN, namely:

$$\mathbf{F} = \sigma(\mathbf{A} * \mathbf{X} * \mathbf{W}), \tag{9}$$

where $\mathbf{W}$ is a learnable weight tensor and $\mathbf{F}$ is the node representation tensor. In TGCN, spatial-temporal features can be directly modeled through tensor face-wise product and M-transformation operations de. For capturing high-order spatial and long-temporal dependences, we can stack multiple TGCN layers to enable further spatial-temporal message passing among nodes. For example, a $L$-layer TGCN formulation can be presented in the form:

$$\mathbf{F} = \sigma\left(\mathbf{A} * \sigma\left(...\sigma(\mathbf{A} * \sigma(\mathbf{A} * \mathbf{X} * \mathbf{W}_1) * \mathbf{W}_2)...\right) * \mathbf{W}_L\right). \tag{10}$$

For TGCN, how to set the mixing matrix M that defines the M-product in (4) is a key issue, which is adopted to aggregate the current and past temporal information of DG. For adopting tensor M-product to operate convolution on the temporal information, we make M lower triangular and banded so that each frontal slice of $\mathbf{X}$ is a linear combination, where we refer to $b$ as the window of M. This choice ensures that each frontal slice of tensor $\mathbf{X}$ only contains information from current to past graph snapshots that are close temporally.

In our model, to better capture the dependence scales between graph snapshots, we develop a dynamic learnable matrix M to learn the mixing values automatically and effectively. Compared to assigning the mixing values manually under certain assumptions, this method not only works more flexibly without such assumptions but also makes mixing values be learned from data. Each mixing value of M is set as:

$$\mathbf{M}_{t,k} = \begin{cases} m_{tk} & \text{if } max(1, t-b+1) \leq k \leq t, \\ 0 & \text{otherwise.} \end{cases} \tag{11}$$

where $m_{tk}$ is the learnable mixing value which can indicate the correlation of the $k$-th temporal information to the $t$-th temporal information. The function $max(i, j)$ returns the maximum value between $i$ and $j$. We normalize each $t$-th mixing vector in M as:

$$m_{tk} \leftarrow \frac{\exp(m_{tk})}{\sum_{j=max(1, t-b+1)}^{t} \exp(\mathbf{M}_{t,j})}, \tag{12}$$

which promises the mixing vector conforming to a probability distribution (i.e., $\sum_{j=max(1, t-b+1)}^{t} \mathbf{M}_{t,j} = 1$).

We capture the spatial-temporal dynamics through $L$-layer TGCN and obtain the node representation tensor $\mathbf{F}$. For the exiting link weight between node $i$ and $j$ at time slot $t$ (i.e. $\mathbf{Y}_{i,j,t}$), we use the learned representation features $\mathbf{F}_{i,:,t}$ and $\mathbf{F}_{j,:,t}$ to predict the value as:

$$\mathbf{Y}_{i,j,t} = \varphi(\mathbf{F}_{i,:,t} \odot \mathbf{F}_{j,:,t} + [\mathbf{F}_{i,:,t} | \mathbf{F}_{j,:,t}]\mathbf{W}_c + \mathbf{z})\mathbf{v}^\top, \tag{13}$$

where $\widetilde{\mathbf{Y}}_{i,j,t}$ is the estimation of $\mathbf{Y}_{i,j,t}$, $W_c$ is the learnable parameter matrix, z is the bias vector and v is regression learnable vector. $\varphi$ is the tanh activation function, $\odot$ represents the element-wise Hadamard product, and $[\cdot \mid \cdot]$ denotes the vector concatenation operation.

*C. Model Optimization*

Since we aim to verify the ability of DG representation in predicting the link weight value task, we chose the regression-based loss for optimizing model parameters. Regression-based loss assumes that estimated results are expected to be equal to the real values. Then, we follow the data density-oriented modeling principle [29-34, 45-50] to build a learning objective defined on known data only. With the commonly adopted Huber loss, the object function is defined as:

$$\ell(\theta) = \begin{cases} \frac{1}{2} \sum_{\mathbf{Y}_{i,j,t} \in \Lambda} \left(\mathbf{Y}_{i,j,t} - \mathbf{Y}_{i,j,t}\right)^2 & \text{if } |\mathbf{Y}_{i,j,t} - \mathbf{Y}_{i,j,t}|_{abs} < \delta, \\ \sum_{\mathbf{Y}_{i,j,t} \in \Lambda} \delta\left(|\mathbf{Y}_{i,j,t} - \mathbf{Y}_{i,j,t}|_{abs} - \frac{\delta}{2}\right) & \text{otherwise.} \end{cases} \quad (14)$$

where $\theta = \{W_n, W, W_c, z, v\}$ is the parameter set to optimize in the proposed TGCN, $\Lambda$ is the training set, and $|\cdot|_{abs}$ computes a number's absolute value. $\delta$ specifies the threshold at which to change between $\delta$-scaled $L_1$ and $L_2$ loss. The value of $\delta$ must be positive and set to 1 in this study.

## IV. EXPERIMENT AND RESULTS

In this section, we present three DG datasets to evaluate the proposed model TGCN and compare it with existing state-of-the-art models.

TABLE I Data Statics

| Dataset | Nodes | Edges | T |
|---|---|---|---|
| D1 | 7,604 | 24,186 | 137 |
| D2 | 16,152 | 159,157 | 174 |
| D3 | 549 | 1,432 | 493 |

*A. General Settings*

**Datasets.** The experiments are conducted on three different DG datasets to evaluate our model TGCN. The detailed statistical information of the datasets is shown in Table I. D1 is a DG of trust changes among Bitcoin users. D2 is composed of dynamic Ethereum transaction data and D3 is collected from personal dynamic terminal communication data.

We split the three datasets with a ratio of 6:1:3 into a training set ($\Lambda$), validation set ($\Omega$), and test set ($\Psi$). In the experiments, we use $\Lambda$ to train the proposed model for optimizing the parameters and then verify the representation ability of the model on $\Omega$ to tune the hyperparameters and select the optimal model after training. Finally, we evaluate the selected optimal model performance on $\Psi$.

**Evaluation Protocol.** This article concerns DG representation learning for the issue of link weight value estimation. Hence, we adopt the estimation accuracy as the evaluation protocol. Mean Absolute Error (MAE) and Root Mean-Squared Error (RMSE) are commonly adopted to measure the estimation error of a tested model [35-39, 50]. They are computed as:

$$\text{MAE} = \left(\sum_{\mathbf{Y}_{i,j,t} \in \Psi} |\mathbf{Y}_{i,j,t} - \mathbf{Y}_{i,j,t}|_{abs}\right) \Big/ |\Psi|,$$

$$\text{RMSE} = \sqrt{\left(\sum_{\mathbf{Y}_{i,j,t} \in \Psi} \left(\mathbf{Y}_{i,j,t} - \mathbf{Y}_{i,j,t}\right)^2\right) \Big/ |\Psi|},$$

where $|\cdot|_{abs}$ returns a number's absolute value and $\Psi$ indicates the test set whose number of entries is denoted by $|\Psi|$. The lower MAE and RMSE values mean higher accuracy in estimating the link weight value of the test model.

**Compared Models.** To evaluate the proposed model, we compare it with eight baseline graph representation learning models, which include two latent feature models (TimeSVD++ [6] (M1) and NeuMF [18] (M2)), two static graph convolution models (GCN [19] (M3) and Light-GCN [20] (M4)), two dynamic graph convolution models (WD-GCN [26] (M5) and Evolve-GCN [27] (M6)). The proposed TGCN is denoted by M7.

*B. Comparison with State-of-the-Art Models*

To validate the representation effectiveness of the proposed TGCN, we compare it with peer models for link weight value estimation, including two models based on latent features, two static GCN models, and two dynamic GCN models. Table II shows the RMSE and MAE obtained by all the above models on the testing set $\Psi$. From the results, we can have the following findings.

a) Compared with two latent feature models (M1 and M2), M7 obtains excellent performance in node representation learning to predict missing link weight. For example, according to Table II on D1, M7's MAE is 34.43%, and 28.44% lower than the M1 and M2, respectively. When considering the evaluation metric RMSE on D1, M7 outperforms M1 and M2 by 16.69% and 15.49%. Similar outcomes are obtained in the other datasets as shown in Table II. The main reason is that M1 and M2 neglect the graph structure of the DG, leading to inferior representation learning performance.

b) M7 demonstrates superior performance in representation learning compared to the static GCN models M3 and M4. According to Table II on D1, M7's MAE is 21.34% and 17.13% lower than the M3 and M4, respectively. In terms of RMSE on D1, M7 outperforms M3 and M4 by 5.22% and 5.64%. Consequently, the inclusion of temporal dependence can significantly enhance representation learning for DGs in comparison to static GCN.

c) M7 also obtains notable performance gain compared with dynamic GCN models M5 and M6, which are models adopting a hybrid archenteric to learn the spatial-temporal dependencies. For instance, as shown in Table II on D1, M7's MAE is 1.4211, which is 9.79% and 6.38% lower than M5 and M6. On the evaluation metric RMSE, M7 obtains 2.7291, which is also better than M5 and M6 by 4.81% and 4.56%. Similar results can be found in other datasets. Therefore, we can conclude that modeling spatial-temporal features under one convolutional framework based on the tensor product is effective for DG representation learning.

TABLE II. RMSE, MAE of M1-M7 on D1-D3.

| No. | Case | D1 | D2 | D3 |
|---|---|---|---|---|
| **M1** | MAE | $2.1672_{\pm 0.0689}$ | $10.4692_{\pm 0.0371}$ | $6.0823_{\pm 0.4421}$ |
|  | RMSE | $3.2757_{\pm 0.0621}$ | $33.6783_{\pm 0.2612}$ | $12.7562_{\pm 0.2185}$ |
| **M2** | MAE | $1.9858_{\pm 0.0114}$ | $10.3863_{\pm 0.0382}$ | $3.5031_{\pm 0.0162}$ |
|  | RMSE | $3.2292_{\pm 0.0363}$ | $25.3318_{\pm 0.1170}$ | $11.0789_{\pm 0.1074}$ |
| **M3** | MAE | $1.8067_{\pm 0.0451}$ | $9.8878_{\pm 0.1191}$ | $2.6333_{\pm 0.0108}$ |
|  | RMSE | $2.8795_{\pm 0.0407}$ | $24.7970_{\pm 0.4206}$ | $10.8574_{\pm 0.0231}$ |
| **M4** | MAE | $1.7148_{\pm 0.0086}$ | $9.6157_{\pm 0.0407}$ | $2.7859_{\pm 0.0034}$ |
|  | RMSE | $2.8922_{\pm 0.0036}$ | $24.1818_{\pm 0.0649}$ | $10.9764_{\pm 0.0021}$ |
| **M5** | MAE | $1.5753_{\pm 0.0839}$ | $9.2589_{\pm 0.0934}$ | $2.6203_{\pm 0.0456}$ |
|  | RMSE | $2.8670_{\pm 0.0695}$ | $23.6486_{\pm 0.0805}$ | $10.4947_{\pm 0.0102}$ |
| **M6** | MAE | $1.5180_{\pm 0.0226}$ | $9.2347_{\pm 0.0944}$ | $2.5687_{\pm 0.0472}$ |
|  | RMSE | $2.8595_{\pm 0.0036}$ | $24.1314_{\pm 0.0780}$ | $10.6712_{\pm 0.0226}$ |
| **M7** | MAE | $\mathbf{1.4211}_{\pm 0.0316}$ | $\mathbf{6.6239}_{\pm 0.0305}$ | $\mathbf{2.2038}_{\pm 0.0151}$ |
|  | RMSE | $\mathbf{2.7291}_{\pm 0.0174}$ | $\mathbf{22.3566}_{\pm 0.0294}$ | $\mathbf{10.4817}_{\pm 0.0256}$ |

## V. CONCLUSION

In this paper, we proposed a tensor graph convolution network for DG representation learning. We first presented information on the DG by tensor form and embedded the node one-hot tensor to latent feature tensor for following tensor convolution calculations. Then, a tensor graph convolutional network was adopted to model spatial-temporal dependencies in one convolutional framework, in which a learnable mixing matrix was designed to learn the temporal dependence scales from data. Extensive experiments on real-world DG datasets were conducted to validate the representation ability of TGCN. The results demonstrated that TGCN outperforms state-of-the-art approaches to obtain DG representations for link weight value estimation.